\documentclass[letterpaper, 10 pt, conference]{ieeeconf}  

\IEEEoverridecommandlockouts                              

\makeatletter
\let\NAT@parse\undefined
\makeatother
\usepackage[comma,numbers, compress]{natbib}

\usepackage[hidelinks]{hyperref}
\usepackage{url}
\usepackage{blindtext}
\usepackage{graphicx}
\usepackage[usenames,dvipsnames]{xcolor}
\usepackage[draft]{fixme}
\fxsetup{theme=color}
\usepackage{soul}

\usepackage{amsmath}
\usepackage{amssymb}
\usepackage{booktabs}
\usepackage{soul}
\usepackage{textcomp}
\usepackage{threeparttable}

\usepackage[capitalize]{cleveref}
\crefname{section}{Sec.}{Sec.}
\crefname{table}{Tab.}{Tab.}

\usepackage{tikz}
\usepackage{pgfplots}
\usepgfplotslibrary{fillbetween}
\usepgfplotslibrary{groupplots}
\usetikzlibrary{arrows}
\usetikzlibrary{positioning,calc}
\usetikzlibrary{decorations.pathreplacing}
\usetikzlibrary{decorations.markings}
\usetikzlibrary{fit}
\usetikzlibrary{shapes.callouts}
\usetikzlibrary{shapes.geometric}
\usetikzlibrary{matrix}
\usetikzlibrary{patterns}

\usepackage[firstpage=true]{background}
\newcommand\copyrighttext{%
        \parbox{\textwidth}{
                \footnotesize
                \textbf{Accepted final version.} IEEE International Conference on Robotics and Automation (ICRA), Montreal, Canada, to appear May 2019
        }
}

\SetBgContents{\copyrighttext}
\SetBgScale{1}
\SetBgColor{black}
\SetBgAngle{0}
\SetBgOpacity{1}
\SetBgPosition{current page.north}
\SetBgVshift{-0.8cm}

\usepackage[bottom]{footmisc}

\newcommand{\etal}{et~al.}

\newcommand{\eg}{e.g.,\ }

\graphicspath{{figures/}}
\title{\LARGE \bf
Towards Learning Abstract Representations for\\Locomotion Planning in High-dimensional State Spaces
}

\author{Tobias Klamt and Sven Behnke
\thanks{All authors are with University of Bonn, Computer Science Institute VI, 
		Autonomous Intelligent Systems, Bonn, Germany
        {\tt\small klamt@ais.uni-bonn.de, behnke@cs.uni-bonn.de}. This work was supported by the European Union's Horizon 2020 Programme under 
        Grant Agreement 644839 (CENTAURO).}%
}

\begin{document}

\maketitle
\thispagestyle{empty}
\pagestyle{empty}

\begin{abstract}

Ground robots which are able to navigate a variety of terrains are needed in many domains.
One of the key aspects is the capability to adapt to the ground structure, which can be realized through movable body parts coming along with additional degrees of freedom (DoF). 
However, planning respective locomotion is challenging since suitable representations result in large state spaces.
Employing an additional abstract representation---which is coarser, lower-dimensional, and semantically enriched---can support the planning.

While a desired robot representation and action set of such an abstract representation can be easily defined, the cost function requires large tuning efforts.  
We propose a method to represent the cost function as a CNN.
Training of the network is done on generated artificial data, while it generalizes well to the abstraction of real world scenes.
We further apply our method to the problem of search-based planning of hybrid driving-stepping locomotion.
The abstract representation is used as a powerful informed heuristic which accelerates planning by multiple orders of magnitude.

\end{abstract}


\section{Introduction}
\label{sec:intro}

Most robot locomotion planners feature search- or sampling-based methods which perform well in 2D and 3D planning problems, \eg driving robot locomotion~\cite{lavalle1998rapidly, kavraki1994probabilistic, hart1968formal}. 
However, many areas of operation, \eg in search and rescue missions, have challenging properties. 
Suitable robots need to adapt to those environments to provide fast, safe, and energy efficient locomotion which can be realized through, \eg tracked flippers, legs, or wheels at adjustable limbs. 
Including these capabilities in the planning problem results in high-dimensional state spaces which may lead to extensive searches for search- and sampling-based approaches.

In recent years, intensive research has been performed on learning-based motion planning approaches~\cite{levine2016end, bojarski2016end, tamar2016value, karkus2017qmdp, srinivas2018universal} which learn to map a given situation to a reasonable action without performing extensive searches.
However, the necessary amount of training data and the required network complexity strongly depend on the size of the considered maps and the number of state space dimensions.
Thus, at the current state-of-the-art, learning-based planners are restricted to small maps or low-dimensional planning problems.

\begin{figure}
	\input{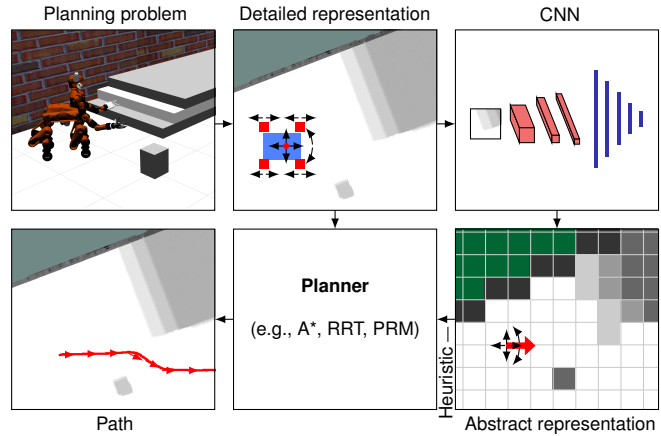}
	\vspace{-0.75cm}
	\caption{A CNN is used to generate an abstract representation of a detailed planning problem which is employed to support the planner.}
	\vspace{-0.5cm}
	\label{fig:teaser}
\end{figure}

A well investigated idea to accelerate planning in large state spaces is abstraction. 
An abstract representation has a coarser resolution or fewer state space dimensions and compensates this information loss through additional features to increase semantics.
Given such an abstract representation, the necessity of the detailed representation during planning may be reduced to certain situations. 
However, a valuable property of an abstract representation to be used along with a detailed representation (\eg as a heuristic or for coarse-to-fine planning) is that the same actions induce the same costs in both representations (\emph{cost similarity}).

In~\cite{klamt2017anytime}, we have proposed a high-dimensional search-based planner for hybrid driving-stepping locomotion which we have extended in~\cite{klamt2018planning} to multiple levels of abstraction. 
This has shown promising results since planning has been significantly accelerated while the resulting path quality has stayed comparable.
However, abstract representations were manually designed and parametrized to obtain \emph{cost similarity} which is a challenging and exhausting task.

In this paper, we propose a method to support the implementation of abstract representations through convolutional neural networks (CNNs), which map a spatially small planning task to a corresponding cost assessment for the shortest path. 
Since these small planning tasks represent a coarse, low-dimensional set of actions, the CNN represents a \emph{cost similar} abstract representation for the high-dimensional planning cost function (see~\cref{fig:teaser}). 

We train the CNN on generated artificial data and evaluate it on simulated and real-world sensor data. 
Furthermore, the method is used to generate a powerful heuristic for hybrid driving-stepping locomotion planning, but it can be easily transferred to other domains, \eg walking locomotion.
The results indicate that the proposed method outperforms our manually tuned approach in terms of abstraction quality, while eliminating tuning efforts, and that the proposed heuristic accelerates path planning by multiple orders of magnitude, compared to popular heuristics.


\section{Related Work}
\label{sec:related_work}

Most robot motion planning approaches are either sampling-based, such as Rapidly-exploring Random Trees (RRT)~\cite{lavalle1998rapidly} or Probabilistic Roadmaps (PRM)~\cite{kavraki1994probabilistic}, search-based, such as A*~\cite{hart1968formal} or a combination of those~\cite{li2014rrt}.
Low-dimensional motion planning in 2D or 3D state spaces, can be seen as solved with these approaches. 
However, it is still challenging to solve high-dimensional, large planning problems since the required computational power and memory significantly increase with an increasing state space size.

A solution to handle large environment sizes is multi-resolution planning~\cite{behnke2003local}.
To handle high-dimensional state spaces, a local adaptation of the robot representation is an option.
In previous work~\cite{klamt2017anytime}, we have proposed a search-based approach to plan hybrid driving-stepping locomotion.
Similarly, Dornbush et al.~\cite{dornbush2018a} have planned multi-modal paths for a humanoid with a search-based planner.
Both approaches handle the occurring high-dimensional state spaces by separating the planning problem with respect to the locomotion mode and apply high-dimensional planning only if required.
Nevertheless, both works suffer the problem of handling large scenarios in feasible time since the high-dimensional represented areas are still too large.

However, those approaches only neglect information in their coarse/low-dimensional representations which might result in wrong assessments, especially for complex terrain.
This is addressed by abstraction: Representations are coarser but semantically enriched to compensate the information loss.
A theoretical basis for abstraction for search-based planning has been given by Holte \etal~\cite{holte1995hierarchical}.
In~\cite{klamt2018planning}, we have extended hybrid driving-stepping locomotion planning to three levels of abstraction.
With increasing abstraction, the environment is represented in a coarser resolution but with additional hand-crafted features such as height differences or terrain classes.
In addition, the robot representation has a coarser resolution and less dimensions with increasing abstraction.
The costs functions were manually tuned to obtain \emph{cost similarity}.
This was done by iteratively comparing costs on a small set of exemplary tasks and adjusting parameters.
The abstract representations accelerate planning by multiple orders of magnitude while the path quality stays comparable.
Especially the utilization of the most abstract representation as a heuristic leads to significant speedup.
However, the design of descriptive features and tuning of cost functions require extensive manual parametrization and are very dependent on the used set of exemplary tasks.

In recent years, learning-based approaches for solving robot motion planning problems have been proposed.
In~\cite{levine2016end} and~\cite{bojarski2016end}, CNNs have been trained to map camera images directly to motor commands, \eg for manipulation tasks or steering of a self-driving car. 
However, the long-term goal-directed behavior of such approaches is usually poor or the training would require unreasonable amounts of data and time.
Tamar et al.~\cite{tamar2016value} have proposed a differentiable approximation of the value iteration algorithm which can be represented as a CNN---the Value Iteration Networks. 
Their performance has been evaluated on small 2D grid worlds.
Similarly, Karkus \etal~\cite{karkus2017qmdp} have proposed QMDP-Net which is also capable of planning in 2D grid worlds.  
Srinivas et al.~\cite{srinivas2018universal} have proposed Universal Planning Networks which map images of the initial and goal scene to actions.
These three approaches point out the general problem of learning-based approaches at the current state-of-the-art: The required amount of training data and the required network complexity are not manageable for large, high-dimensional planning problems.

To summarize, learning-based planning approaches can handle local problems with limited state space sizes quickly without performing extensive searches. 
In contrast, traditional planning approaches show good goal-directed behavior but might get stuck in extensive searches for complex high-dimensional problems.
Hence, it promising to combine these approaches and merge the advantages of both.
Faust \etal~\cite{faust2018prmrl} use a reinforcement learning agent to learn short-range, point-to-point navigation policies for 2D and 3D action spaces which capture the robot dynamic and task constraint without considering the large-scale topology. 
Sampling-based planning is used to plan waypoints which give the planning a long-range goal-directed behavior.

In contrast to that work, we combine learning- and search-based planning to handle 7-dimensional hybrid driving-stepping locomotion planning.
A CNN represents the cost function of an abstract representation of the high-dimensional planning problem employed as a heuristic to accelerate planning.


\section{Problem Statement}

\begin{figure*}
	\input{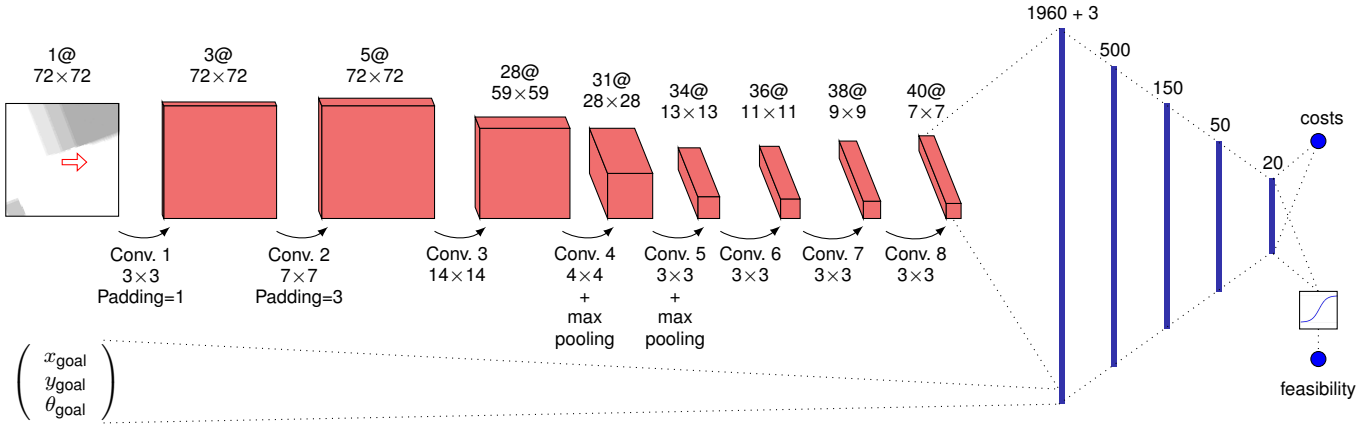}
	\vspace{-0.4cm}
	\caption{Architecture of the proposed CNN. Input are a height map patch and the goal state. Although it is not fed into the network, the start state is depicted as a red arrow for better understanding. Output are the \emph{feasibility} and \emph{costs} values. Convolutional layers are visualized as red cuboids; blue lines show fully connected layers. If not stated different, convolutions have a padding of 0 and a stride of 1.}
	\label{fig:network_architecture}	
	\vspace{-0.3cm}
\end{figure*}

Given a planner which uses an environment representation ($\mathcal{E}$), a robot representation ($\mathcal{R}_\text{d}$), a corresponding action set ($\mathcal{A}_\text{d}$), and a cost function ($\mathcal{C}_\text{d}$). 
$\mathcal{E}$ is a map with an arbitrary number of features describing each cell.
$\mathcal{R}_\text{d}$ represents all required DoF of the robot kinematics which are necessary to address the planning problem.
$\mathcal{A}_\text{d}$ contains all actions which can be executed by the robot such that $r_\text{d,i} + a_\text{d,j}  = r_\text{d,i+1}$,
an action $a_\text{d,j} \in \mathcal{A}_\text{d}$ connects two successive robot states $r_\text{d,i}$, $r_\text{d,i+1} \in \mathcal{R}_\text{d}$ while inducing the costs $\mathcal{C}_\text{d}(r_\text{d,i}, a_\text{d,j})$.

A second, abstract representation, consisting of $\mathcal{E}$, $\mathcal{R}_\text{a}$, $\mathcal{A}_\text{a}$, and $\mathcal{C}_\text{a}$, can be used to support the planning.
$\mathcal{R}_\text{a}$ describes the robot state in a low-dimensional state space although this might not suffice to describe the robot state in enough detail for execution. 
The correspondence between an abstract robot state $r_\text{a,i} \in \mathcal{R}_\text{a}$ and a detailed robot state $r_\text{d,i} \in \mathcal{R}_\text{d}$ is given through the transformation 
\begin{equation}
	r_\text{a,i} = \mathcal{T}_\text{d $\mapsto$ a}(r_\text{d,i})
\end{equation}
and vice versa with
\begin{equation}
	r_\text{d,i} = \mathcal{T}_\text{a $\mapsto$ d}(r_\text{a,i}) \text{.}
\end{equation}

$\mathcal{A}_\text{a}$ describes actions $a_\text{a,j}$ to move an abstract robot state \mbox{$r_\text{a,i} \in \mathcal{R}_\text{a}$} to a successive state $r_\text{a,i+1} \in \mathcal{R}_\text{a}$.
The resolution of $\mathcal{A}_\text{a}$ is coarser compared to $\mathcal{A}_\text{d}$, such that an action sequence 
\begin{equation}
	\mathcal{T}_\text{a $\mapsto$ d}(r_\text{a,i}) + a_\text{d,j} + a_\text{d,j+1} + ... + a_\text{d,j+k} = \mathcal{T}_\text{a $\mapsto$ d}(r_\text{a,i+1}) \text{,}
\end{equation}
is necessary in the detailed representation to perform the least cost transition between two successive robot states in the abstract representation, while the difference between the abstract costs $\mathcal{C}_\text{a}(r_\text{a,i}, a_\text{a,j})$ and the detailed costs $\mathcal{C}_\text{d}(\mathcal{T}_\text{a $\mapsto$ d}(r_\text{a,i}), a_\text{d,j}, ..., a_\text{d,j+k})$ should be minimized to obtain \emph{cost similarity}.

While $\mathcal{R}_\text{a}$ and $\mathcal{A}_\text{a}$ can be easily defined, $\mathcal{C}_\text{a}$ needs extensive tuning. 
We propose to represent $\mathcal{C}_\text{a}$ as a CNN to avoid these tuning efforts and improve abstraction quality.


\section{Network Design}

We propose a regular CNN architecture---consisting of convolutional layers and successive fully connected layers---to learn the abstract cost function (see~\cref{fig:network_architecture}).
Input are a height map patch with 72\,$\times$\,72 pixels and the three-dimensional abstract goal state $r_\text{a,g}$.
The start state $r_\text{a,s}$ is assumed to be always in the map patch center with a fixed orientation and is not fed into the network.
For a given resolution of 2.5\,cm, the map size is chosen such that for every abstract goal state $r_\text{a,g}=r_\text{a,s} + a_\text{a,j} \in \mathcal{A}_\text{a}$, the corresponding detailed goal state $r_\text{d,g} = \mathcal{T} _\text{a $\mapsto$ d}(r_\text{a,g})$ with any feasible leg configuration is completely inside this map patch.
$r_\text{a,g}$ is defined in resolution steps relative to $r_\text{a,s}$.

Instead of only outputting costs which become infinite for infeasible queries, we output two values:
The \emph{feasibility} value describes whether there exists a path between $r_\text{a,s}$ and $r_\text{a,g}$, and,
if so, the \emph{costs} value describes the corresponding costs.

We discovered that key to a good abstraction performance are some convolutions with large kernel sizes which might be explained as follows:
A first small convolution extracts descriptive map features from the input height map.
Next, a second convolution possesses a kernel size which is similar to the size of a robot foot and thus can determine if foot placement is possible for each kernel position.
The kernel size of the third convolution is chosen such that is covers the maximum action length for an individual foot.
Hence, it can find connections between feasible foot positions in a certain distance which is valuable for, \eg steps.
The following convolutions and max pooling operations with small kernels do further processing on the actions and are followed by six fully connected layers.

The last fully connected layer is split:
While \emph{costs} are output directly, the \emph{feasibility} output is processed by a sigmoid function since it is Boolean.


\section{Learning Abstraction of Hybrid Driving-Stepping Locomotion Planning}
\label{sec:learning_hybrid_locomotion}

\begin{figure}
	\input{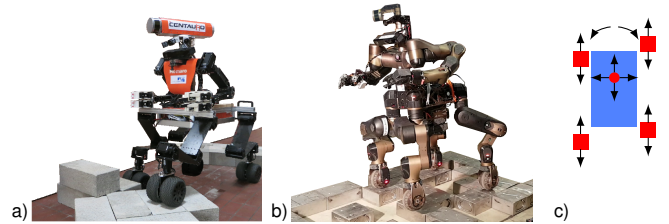}
	\vspace{-0.6cm}
	\caption{Hybrid driving-stepping locomotion robots. a) Momaro, b) Centauro, c) corresponding detailed robot representation (blue = robot base, red squares = feet, arrows visualize the DoF).}
	\label{fig:robot_overview}
	\vspace{-0.4cm}
\end{figure}

We apply the proposed method to hybrid driving-stepping locomotion planning for our platforms \eg Momaro~\cite{schwarz2017nimbro} and Centauro~\cite{klamt2018supervised} (see~\cref{fig:robot_overview}\,a,\,b). 
Both are able to perform omnidirectional driving and stepping motions. 
A detailed robot representation which matches both robots is depicted in~\cref{fig:robot_overview}\,c. 

\subsection{Detailed Representation}

\begin{figure}
	\centering
	\input{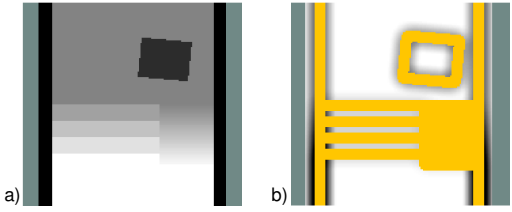}
	\vspace{-0.3cm}
	\caption{Detailed environment representation: a) input height map, b) foot cost map (yellow = untraversable by driving, olive = unknown).}
	\label{fig:detailed_environment}
\end{figure}

The environment representation $\mathcal{E}$ is a height map which is generated from registered point clouds using the method by Droeschel \etal~\citep{Droeschel2017104} (\cref{fig:detailed_environment}\,a).
Regarding $\mathcal{C}_\text{d}$, foot costs (see~\cref{fig:detailed_environment}\,b) and base costs, which describe the costs to place a single foot/the base at a given position/in a given state on the map, are computed from this height map.
Foot costs and base costs are merged to state costs.
The robot is represented in 7D states $r_\text{d} \in \mathcal{R}_\text{d}=(r_\text{x}, r_\text{y}, r_\theta, f_\text{1}, ... , f_\text{4})$ with the robot base state ($r_\text{x}, r_\text{y}, r_\theta$) and the relative longitudinal position of each foot $f_\text{1}, ... , f_\text{4}$ as shown in~\cref{fig:robot_overview}\,c.
Positions have a resolution of 2.5\,cm while there are 64 discrete orientations. 
Lateral foot positions are fixed and foot heights are computed after a result path is found.

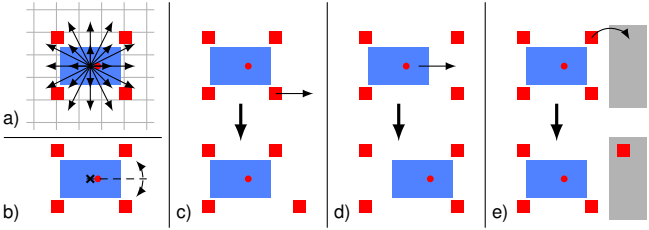
\begin{figure}
	\begin{tikzpicture}[
 	font=\sffamily\footnotesize,
    every node/.append style={text depth=.2ex},
	l/.style={font=\sffamily\scriptsize},
]


\draw (2.1,0) -- ++(0, 3);
\draw (4.2,0) -- ++(0, 3);
\draw (6.3,0) -- ++(0, 3);

\definecolor {robot_base}{RGB}{80, 130, 255};

\draw[black!30] (1.8,1.3) -- ++(0, 1.7);
\draw[black!30] (1.5,1.3) -- ++(0, 1.7);
\draw[black!30] (1.2,1.3) -- ++(0, 1.7);
\draw[black!30] (0.9,1.3) -- ++(0, 1.7);
\draw[black!30] (0.6,1.3) -- ++(0, 1.7);
\draw[black!30] (0.3,1.3) -- ++(0, 1.7);

\draw[black!30] (0.2,1.4) -- ++(1.7, 0);
\draw[black!30] (0.2,1.7) -- ++(1.7, 0);
\draw[black!30] (0.2,2.0) -- ++(1.7, 0);
\draw[black!30] (0.2,2.3) -- ++(1.7, 0);
\draw[black!30] (0.2,2.6) -- ++(1.7, 0);
\draw[black!30] (0.2,2.9) -- ++(1.7, 0);

\coordinate (robot_center) at (1.05, 2.15);
\fill[robot_base] ($(robot_center) +(-0.4,-0.25)$) rectangle ++(0.8,0.5);
\fill[red] ($(robot_center) +(0.38,0.29)$) rectangle ++(0.17,0.17);
\fill[red] ($(robot_center) +(0.38,-0.45)$) rectangle ++(0.17,0.17);
\fill[red] ($(robot_center) +(-0.52,0.29)$) rectangle ++(0.17,0.17);
\fill[red] ($(robot_center) +(-0.52,-0.45)$) rectangle ++(0.17,0.17);
\fill[red] ($(robot_center) +(0.1,0)$) circle (0.045);

\draw[-latex](1.05,2.15) -- ++(0.3, 0);
\draw[-latex](1.05,2.15) -- ++(0.0, 0.3);
\draw[-latex](1.05,2.15) -- ++(-0.30, 0);
\draw[-latex](1.05,2.15) -- ++(0, -0.3);
\draw[-latex](1.05,2.15) -- ++(0.3, 0.3);
\draw[-latex](1.05,2.15) -- ++(0.3, -0.3);
\draw[-latex](1.05,2.15) -- ++(-0.3, 0.3);
\draw[-latex](1.05,2.15) -- ++(-0.3, -0.3);
\draw[-latex](1.05,2.15) -- ++(0.6, 0);
\draw[-latex](1.05,2.15) -- ++(0, 0.6);
\draw[-latex](1.05,2.15) -- ++(-0.6, 0);
\draw[-latex](1.05,2.15) -- ++(0, -0.6);
\draw[-latex](1.05,2.15) -- ++(0.3, 0.6);
\draw[-latex](1.05,2.15) -- ++(0.3, -0.6);
\draw[-latex](1.05,2.15) -- ++(-0.3, 0.6);
\draw[-latex](1.05,2.15) -- ++(-0.3, -0.6);
\draw[-latex](1.05,2.15) -- ++(0.6, 0.3);
\draw[-latex](1.05,2.15) -- ++(0.6, -0.3);
\draw[-latex](1.05,2.15) -- ++(-0.6, 0.3);
\draw[-latex](1.05,2.15) -- ++(-0.6, -0.3);

\draw(-0.1,1.2)-- ++(2.1,0);

\node at (0.0,1.45)[l] {a)};

\coordinate (robot_center) at (1.05, 0.65);
\fill[robot_base] ($(robot_center) +(-0.4,-0.25)$) rectangle ++(0.8,0.5);
\fill[red] ($(robot_center) +(0.38,0.29)$) rectangle ++(0.17,0.17);
\fill[red] ($(robot_center) +(0.38,-0.45)$) rectangle ++(0.17,0.17);
\fill[red] ($(robot_center) +(-0.52,0.29)$) rectangle ++(0.17,0.17);
\fill[red] ($(robot_center) +(-0.52,-0.45)$) rectangle ++(0.17,0.17);
\fill[red] ($(robot_center) +(0.1,0)$) circle (0.045);

\draw[-latex](1.75,0.67) arc (5:40:0.4);
\draw[-latex](1.75,0.63) arc (-5:-40:0.4);
\draw[densely dashed](1.0,0.65) -- ++(0.8,0);
\draw[thick](1.0,0.6) -- ++(0.1,0.1);
\draw[thick](1.0,0.7) -- ++(0.1,-0.1);

\node at (0.0, 0.2)[l] {b)};

\coordinate (robot_center) at (3.05, 2.15);
\fill[robot_base] ($(robot_center) +(-0.4,-0.25)$) rectangle ++(0.8,0.5);
\fill[red] ($(robot_center) +(0.38,0.29)$) rectangle ++(0.17,0.17);
\fill[red] ($(robot_center) +(0.38,-0.45)$) rectangle ++(0.17,0.17);
\fill[red] ($(robot_center) +(-0.52,0.29)$) rectangle ++(0.17,0.17);
\fill[red] ($(robot_center) +(-0.52,-0.45)$) rectangle ++(0.17,0.17);
\fill[red] ($(robot_center) +(0.1,0)$) circle (0.045);

\draw[-latex]($(robot_center) +(0.465,-0.365)$) -- ++(0.5,0);
\draw[-latex, very thick]($(robot_center) +(0,-0.5)$) -- ++(0,-0.5);

\coordinate (robot_center) at (3.05, 0.65);
\fill[robot_base] ($(robot_center) +(-0.4,-0.25)$) rectangle ++(0.8,0.5);
\fill[red] ($(robot_center) +(0.38,0.29)$) rectangle ++(0.17,0.17);
\fill[red] ($(robot_center) +(0.7,-0.45)$) rectangle ++(0.17,0.17);
\fill[red] ($(robot_center) +(-0.52,0.29)$) rectangle ++(0.17,0.17);
\fill[red] ($(robot_center) +(-0.52,-0.45)$) rectangle ++(0.17,0.17);
\fill[red] ($(robot_center) +(0.1,0)$) circle (0.045);

\node at (2.3, 0.2)[l] {c)};

\coordinate (robot_center) at (5.15, 2.15);
\fill[robot_base] ($(robot_center) +(-0.4,-0.25)$) rectangle ++(0.8,0.5);
\fill[red] ($(robot_center) +(0.7,0.29)$) rectangle ++(0.17,0.17);
\fill[red] ($(robot_center) +(0.7,-0.45)$) rectangle ++(0.17,0.17);
\fill[red] ($(robot_center) +(-0.52,0.29)$) rectangle ++(0.17,0.17);
\fill[red] ($(robot_center) +(-0.52,-0.45)$) rectangle ++(0.17,0.17);
\fill[red] ($(robot_center) +(0.1,0)$) circle (0.045);

\draw[-latex]($(robot_center) +(0.265,0.0)$) -- ++(0.5,0);
\draw[-latex, very thick]($(robot_center) +(0,-0.5)$) -- ++(0,-0.5);

\coordinate (robot_center) at (5.47, 0.65);
\fill[robot_base] ($(robot_center) +(-0.4,-0.25)$) rectangle ++(0.8,0.5);
\fill[red] ($(robot_center) +(0.38,0.29)$) rectangle ++(0.17,0.17);
\fill[red] ($(robot_center) +(0.38,-0.45)$) rectangle ++(0.17,0.17);
\fill[red] ($(robot_center) +(-0.84,0.29)$) rectangle ++(0.17,0.17);
\fill[red] ($(robot_center) +(-0.84,-0.45)$) rectangle ++(0.17,0.17);
\fill[red] ($(robot_center) +(0.1,0)$) circle (0.045);

\node at (4.4, 0.2)[l] {d)};

\coordinate (robot_center) at (7.25, 2.15);
\fill[robot_base] ($(robot_center) +(-0.4,-0.25)$) rectangle ++(0.8,0.5);
\fill[red] ($(robot_center) +(0.38,0.29)$) rectangle ++(0.17,0.17);
\fill[red] ($(robot_center) +(0.38,-0.45)$) rectangle ++(0.17,0.17);
\fill[red] ($(robot_center) +(-0.52,0.29)$) rectangle ++(0.17,0.17);
\fill[red] ($(robot_center) +(-0.52,-0.45)$) rectangle ++(0.17,0.17);
\fill[red] ($(robot_center) +(0.1,0)$) circle (0.045);

\fill[black!30]($(robot_center) +(0.7, -0.55)$) rectangle ++(0.5, 1.1);

\draw[-latex]($(robot_center) +(0.465,0.375)$) arc (145:30:0.3);
\draw[-latex, very thick]($(robot_center) +(0,-0.5)$) -- ++(0,-0.5);

\coordinate (robot_center) at (7.25, 0.65);
\fill[black!30]($(robot_center) +(0.7, -0.55)$) rectangle ++(0.5, 1.1);
\fill[robot_base] ($(robot_center) +(-0.4,-0.25)$) rectangle ++(0.8,0.5);
\fill[red] ($(robot_center) +(0.8,0.29)$) rectangle ++(0.17,0.17);
\fill[red] ($(robot_center) +(0.38,-0.45)$) rectangle ++(0.17,0.17);
\fill[red] ($(robot_center) +(-0.52,0.29)$) rectangle ++(0.17,0.17);
\fill[red] ($(robot_center) +(-0.52,-0.45)$) rectangle ++(0.17,0.17);
\fill[red] ($(robot_center) +(0.1,0)$) circle (0.045);

\node at (6.5, 0.2) [l]{e)};

\end{tikzpicture}
	\vspace{-0.7cm}
	\caption{Robot actions in $\mathcal{A}_\text{d}$: a) omnidirectional driving with fixed orientation, b) turning with fixed position, c) moving a foot relative to the base while keeping ground contact, d) longitudinal base shift, e) step. Grid and orientation resolution are enlarged to facilitate visualization.}
	\label{fig:detailed_actions}
	\vspace{-0.3cm}
\end{figure}

Robot actions in $\mathcal{A}_\text{d}$ are (see~\cref{fig:detailed_actions}):
\begin{itemize}
	\item omnidirectional driving within a 20-neighborhood with fixed orientation,
	\item turning to the next discrete orientation,
	\item moving an individual foot relative to the robot base while keeping ground contact,
	\item moving the base longitudinal relative to the feet, and
	\item performing a step with a single foot.
\end{itemize}
Steps are represented as the direct transition from a pre-stepping to a post-stepping state.
Only those steps in the result path are refined to detailed motion sequences which consider robot stability and the detailed stepping motion. 
Each action carries costs with respect to the occurring foot and base costs that the individual robot elements experience.
For driving locomotion, a large angular difference between the robot orientation and driving direction is punished with higher costs to prefer driving forward which brings advantages to the perception of the environment directly in front of the robot and when switching to stepping motions in the sagittal direction.
An A*-based planner which uses the above presented representation is used to plan hybrid driving-stepping locomotion paths.
More details can be found in~\cite{klamt2017anytime}.

\subsection{Abstract Representation}

$\mathcal{R}_\text{a}$ contains 3D robot states $r_\text{a}=(r_\text{x}, r_\text{y}, r_\theta)$ which describe the robot position with a resolution of 10\,cm and the orientation in 16 discrete steps.
Individual foot configurations are neglected.
$\mathcal{A}_\text{a}$ contains 
\begin{itemize}
	\item moving the robot within a 20-neighborhood with fixed orientation (see~\cref{fig:detailed_actions}\,a) and
	\item turning to the next discrete orientation with fixed position (see~\cref{fig:detailed_actions}\,b).
\end{itemize}

Transforming a detailed robot state to an abstract robot state ($\mathcal{T}_\text{d $\mapsto$ a}$) is done by neglecting the foot positions and matching the position and orientation to the coarse resolution of the abstract state space. 
The transformation from an abstract to a detailed robot state $\mathcal{T}_\text{a $\mapsto$ d}$ is more complicated: For all detailed robot base states that match the abstract state, we search the least cost foot configuration while preferring configurations which are close to the neutral robot configuration (\cref{fig:detailed_actions}\,a). 
The detailed robot state with the minimum state costs is the transformation result.

\subsection{Network Training}
\label{sec:train_network}

Training data is generated artificially.
Hence, large datasets can be produced without considerable effort.
A map generator produces height maps of the desired network input size. 
The following obstacles are placed randomly in those height maps:
\begin{itemize}
	\item cuboid shaped obstacles of random size,
	\item walls of random length and height, and
	\item staircases of random width with a random number of stairs (with random height and length). 
\end{itemize}
We produce 2,000 maps of each of the following categories:
\begin{itemize}
	\item one/two/three cuboid obstacles,
	\item one/two walls,
	\item one cuboid obstacle and one wall,
	\item one staircase,
	\item one staircase and one wall, and
	\item one staircase whose orientation is in the interval $\left[-\frac{\pi}{16}, \frac{\pi}{16} \right]$ around the robot orientation. Those maps are used to set a learning focus on stair climbing.
\end{itemize}
For each map, we define 22 abstract goal states $r_\text{a,g$_\text{i}$}$ with respect to $\mathcal{A}_\text{a}$.
The start state $r_\text{a,s}$ is always in the map center with a fixed orientation.
$r_\text{a,s}$ and $r_\text{a,g$_\text{i}$}$ are transformed to $\mathcal{R}_\text{d}$ using $\mathcal{T}_\text{a $\mapsto$ d}$.
For some maps, a valid detailed start state $r_\text{d,s}$ cannot be found due to obstacles.
Those maps are deleted.
In total, we get a set of 11,327 maps with 249,194 tasks.
Subsequently, we search for a shortest path from $r_\text{d,s}$ to $r_\text{d,g$_\text{i}$}$ with our detailed A*-planner.
For each task, we save the \emph{feasibility} flag which describes if a path could be found. 
\emph{Costs} are saved for all feasible tasks.

\begin{figure}
	\centering
	\begin{tikzpicture}[
 	font=\sffamily\footnotesize,
    every node/.append style={text depth=.2ex},
	l/.style={font=\sffamily\scriptsize},
    b/.style={font=\bf},
]



\pgfplotsset{every tick label/.append style={font=\sffamily\scriptsize}, every label/.append style={font=\sffamily\scriptsize}}

\begin{axis}[
			xmin=1, xmax=100,
			ymin=0, ymax=100,
			no marks,
			xlabel={Epochs},
			ylabel={Feasibility correct [\%]},
			width=0.9*\linewidth,
			height=0.7*\linewidth,
			xlabel near ticks,
			ylabel near ticks,
			axis y line*=left,
]
\addplot coordinates{(1,89.95) 
									   (2,93.12)
									   (3,94.35)
									   (4,94.64)
									   (5,93.13)
									   (6,95.01)
									   (7,95.04)
									   (8,94.82)
									   (9,94.98)
									 (10,95.25)
									 (11,92.64)
									 (12,95.11)
									 (13,95.34)
									 (14,95.44)
									 (15,95.51)
									 (16,95.17)
									 (17,95.22)
					(18,95.80)
(19,95.68)
(20,95.64)
(21,95.70)
(22,95.62)
(23,95.66)
(24,95.58)
(25,95.69)
(26,95.58)
(27,95.73)
(28,95.68)
(29,95.55)
(30,95.62)
(31,95.72)
(32,95.56)
(33,95.58)
(34,95.67)
(35,95.57)
(36,95.78)
(37,95.57)
(38,95.63)
(39,95.58)
(40,95.53)
(41,95.55)
(42,95.62)
(43,95.41)
(44,95.56)
(45,95.46)
(46,95.49)
(47,95.46)
(48,95.45)
(49,95.43)
(50,95.42)
(51,95.38)
(52,95.22)
(53,95.40)
(54,95.37)
(55,95.35)
(56,95.27)
(57,95.36)
(58,95.29)
(59,95.34)
(60,95.30)
(61,95.23)
(62,95.26)
(63,95.26)
(64,95.29)
(65,95.24)
(66,95.29)
(67,95.27)
(68,95.29)
(69,95.36)
(70,95.23)
(71,95.24)
(72,95.26)
(73,95.18)
(74,95.10)
(75,95.17)
(76,95.24)
(77,95.24)
(78,95.08)
(79,95.09)
(80,95.12)
(81,95.14)
(82,95.07)
(83,95.14)
(84,95.15)
(85,95.00)
(86,95.09)
(87,95.01)
(88,94.93)
(89,94.93)
(90,95.00)
(91,95.00)
(92,95.00)
(93,95.00)
(94,95.00)
(95,95.00)
(96,95.03)
(97,95.00)
(98,95.01)
(99,95.00)
(100,95.00)
};
\label{plot_feas}
\end{axis}

\begin{axis}[ xmin=1, xmax=100,
						ymin=0, ymax=0.8,
						area style,
						enlarge x limits=false,
						width=0.9*\linewidth,
						height=0.7*\linewidth,
						ylabel near ticks,
						axis y line*=right,
						area legend,
						xticklabels={,,},
						ylabel={Costs [ - ]},
] 
\addplot[name path=plot1, draw opacity=0] coordinates {(0,0.258) (100, 0.258)};
\addplot[name path=plot2, draw opacity=0] coordinates {(0,0.689) (100, 0.689)};
\addplot[black!20] fill between[of = plot1 and plot2];
\label{plot_costs_std_dev}
\end{axis}

\begin{axis}[
			xmin=1, xmax=50,
			ymin=0, ymax=0.8,
			mark options={scale=0.3},
			width=0.9*\linewidth,
			height=0.7*\linewidth,
			xlabel near ticks,
			ylabel near ticks,
			axis y line*=right,
			xticklabels={,,},
			yticklabels={,,}
]
\addplot [color = black] coordinates{(1,0.4737) (50,0.4737)
									 
};
\label{plot_costs_base_line}
\end{axis}

\begin{axis}[ xmin=1, xmax=100,
						ymin=0, ymax=0.8,
						l,
						area style,
						enlarge x limits=false,
						width=0.9*\linewidth,
						height=0.7*\linewidth,
						ylabel near ticks,
						axis y line*=right,
						area legend,
						xticklabels={,,},
						yticklabels={,,}
] 

\addplot[red, dashed] coordinates {
(1,0.081) 
(2,0.062) 
(3,0.056) 
(4,0.051) 
(5,0.050) 
(6,0.056) 
(7,0.054) 
(8,0.052) 
(9,0.053) 
(10,0.048) 
(11,0.052) 
(12,0.047) 
(13,0.049) 
(14,0.045) 
(15,0.042) 
(16,0.043) 
(17,0.044) 
(18,0.038) 
(19,0.038) 
(20,0.038) 
(21,0.037) 
(22,0.038) 
(23,0.041) 
(24,0.038) 
(25,0.036) 
(26,0.036) 
(27,0.042) 
(28,0.037) 
(29,0.036) 
(30,0.035) 
(31,0.043) 
(32,0.036) 
(33,0.037) 
(34,0.037) 
(35,0.038) 
(36,0.035) 
(37,0.035) 
(38,0.035) 
(39,0.035) 
(40,0.039) 
(41,0.036) 
(42,0.034) 
(43,0.034) 
(44,0.035) 
(45,0.034) 
(46,0.034) 
(47,0.037) 
(48,0.032) 
(49,0.032) 
(50,0.031)
(51,0.031)
(52,0.031)
(53,0.031)
(54,0.032)
(55,0.032)
(56,0.031)
(57,0.031)
(58,0.030)
(59,0.031)
(60,0.029)
(61,0.029)
(62,0.030)
(63,0.030)
(64,0.030)
(65,0.030)
(66,0.030)
(67,0.030)
(68,0.030)
(69,0.030)
(70,0.029)
(71,0.030)
(72,0.029)
(73,0.030)
(74,0.030)
(75,0.029)
(76,0.029)
(77,0.029)
(78,0.028)
(79,0.028)
(80,0.029)
(81,0.030)
(82,0.029)
(83,0.029)
(84,0.029)
(85,0.030)
(86,0.030)
(87,0.031)
(88,0.029)
(89,0.030)
(90,0.030)
(91,0.030)
(92,0.030)
(93,0.030)
(94,0.030)
(95,0.030)
(96,0.030)
(97,0.030)
(98,0.030)
(99,0.030)
(100,0.030)
};
\label{plot_learned_costs_error}

\end{axis}

\begin{axis}[xmin=1, xmax=100,
						ymin=0, ymax=0.8,
						l,
						ylabel={},
						width=0.9*\linewidth,
						height=0.7*\linewidth,
						axis y line*=right,
						legend pos=south east,
						legend style ={font=\sffamily\scriptsize, at={(0.97,0.07)}},
						xticklabels={,,},
						yticklabels={,,}
]

\addlegendimage{/pgfplots/refstyle=plot_feas}\addlegendentry{Learned feasibility}
\addlegendimage{/pgfplots/refstyle=plot_costs_base_line}\addlegendentry{$\O$ $\mathcal{C}_\text{d}$} 
\addlegendimage{/pgfplots/refstyle=plot_costs_std_dev}\addlegendentry{Std. dev.($\mathcal{C}_\text{d}$)}

\addplot [color=red] coordinates{
(1,0.403)
(2,0.460)
(3,0.437)
(4,0.446)
(5,0.455)
(6,0.463)
(7,0.457)
(8,0.453)
(9,0.456)
(10,0.446)
(11,0.448)
(12,0.427)
(13,0.460)
(14,0.438)
(15,0.447)
(16,0.440)
(17,0.454)
(18,0.453)
(19,0.453)
(20,0.434)
(21,0.439)
(22,0.437)
(23,0.437)
(24,0.443)
(25,0.451)
(26,0.451)
(27,0.468)
(28,0.457)
(29,0.448)
(30,0.449)
(31,0.471)
(32,0.443)
(33,0.459)
(34,0.438)
(35,0.442)
(36,0.442)
(37,0.447)
(38,0.454)
(39,0.447)
(40,0.436)
(41,0.442)
(42,0.454)
(43,0.440)
(44,0.461)
(45,0.457)
(46,0.440)
(47,0.434)
(48,0.452)
(49,0.456)
(50,0.447)
(51,0.449)
(52,0.447)
(53,0.453)
(54,0.460)
(55,0.452)
(56,0.455)
(57,0.457)
(58,0.449)
(59,0.457)
(60,0.452)
(61,0.450)
(62,0.449)
(63,0.458)
(64,0.450)
(65,0.450)
(66,0.449)
(67,0.458)
(68,0.448)
(69,0.458)
(70,0.451)
(71,0.456)
(72,0.456)
(73,0.455)
(74,0.458)
(75,0.452)
(76,0.453)
(77,0.451)
(78,0.454)
(79,0.453)
(80,0.453)
(81,0.450)
(82,0.458)
(83,0.453)
(84,0.456)
(85,0.458)
(86,0.457)
(87,0.448)
(88,0.448)
(89,0.453)
(90,0.456)
(91,0.452)
(92,0.455)
(93,0.455)
(94,0.455)
(95,0.456)
(96,0.457)
(97,0.454)
(98,0.457)
(99,0.455)
(100,0.455)
};
\addlegendentry{$\O$ $\mathcal{C}_\text{a, learned}$};
\addlegendimage{/pgfplots/refstyle=plot_learned_costs_error}\addlegendentry{$\O$ Error($\mathcal{C}_\text{a, learned}$)}
\end{axis}

\fill[black!20](3.16,1.24) rectangle ++(0.6,0.2);

\end{tikzpicture}
	\vspace{-0.3cm}
	\caption{CNN training performance. The detailed cost function $\mathcal{C}_\text{d}$ is shown as a base line. The stated error describes the cost difference between the detailed and the abstract cost functions.}
	\label{fig:learning_performance_graph}
	\vspace{-0.5cm}
\end{figure}
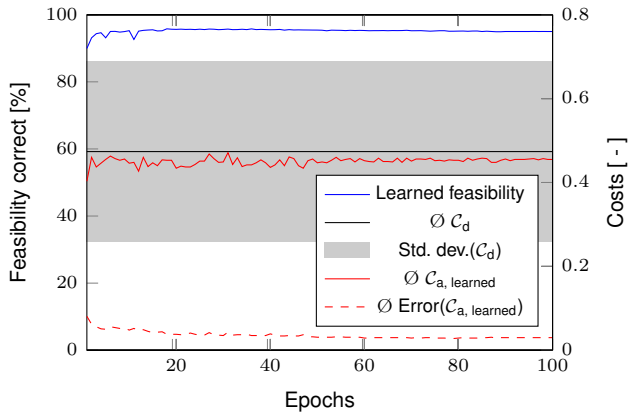

The network is trained using the SGD optimizer with a learning rate of $0.0001$ and a momentum of $0.9$.
We use a BCE loss function for the \emph{feasibility} and a L1 loss function for the \emph{costs}.
The \emph{costs} loss is only considered in the backpropagation if the task is feasible.
Losses are weighted with $\mathcal{W}_\text{feasible}$ and $\mathcal{W}_\text{costs}$, both starting at $1$.
If no improvement by means of a decreasing loss is achieved in three successive training epochs, the corresponding loss weight is divided by $5$.
This dynamic is applied to both losses individually.
For evaluation, a threshold of $0.5$ is used to make the \emph{feasibility} output Boolean.
A validation set which includes 100 maps of each mentioned category is generated and used to evaluate the training performance (\cref{fig:learning_performance_graph}).
We train the network for 100 epochs and choose the state with the best results on the validation set for our experiments.

\subsection{Abstract Representation as Heuristic}
We utilize the learned abstract representation as a heuristic for planning in the detailed representation.
For a given goal state $r_\text{d,g}$, a one-to-any 3D Dijkstra search is started from $r_\text{a,g} = \mathcal{T} _\text{d $\mapsto$ a}(r_\text{d,g})$ and explores the whole map in the abstract representation.
During that search, neighbor states $r_\text{a,n$_\text{i}$}$ for a state $r_\text{a,m}$ are generated through abstract actions $a_\text{a,i}$ such that $r_\text{a, n$_\text{i}$} + a_\text{a,i} = r_\text{a, m}$ while respective costs are computed by the CNN which is fed with the respective height map patches.
Start and goal of each action are exchanged since this search is running backwards, starting at the planner goal state.
While actions which are assessed as infeasible are neglected, feasible actions are assigned the corresponding \emph{costs} output.

Consequently, each abstract state in the map carries the estimated costs of the shortest path to $r_\text{d,g}$.
When planning in the detailed representation, the planner uses these cost estimations as an informed heuristic.

Please note that we cannot prove that this heuristic always underestimates costs, and thus, we cannot prove admissibility for the generation of optimal paths. 
We rather focus on the generation of paths with a satisfying quality in feasible time and thus, accept sub-optimality to speedup planning.

The CNN is implemented using Python 2.7 and PyTorch 0.4.1. The planner is implemented in C++. Communication is realized via ROS. Code for the CNN, the training data generator and the framework to use the CNN as a heuristic is available online\footnote{\url{https://github.com/AIS-Bonn/planning_abstraction_net}}.


\section{Experiments}
\label{sec:experiments}

We evaluate the proposed approach in two experiments which compare the abstraction quality to the manually tuned abstraction of our previous work~\cite{klamt2018planning} and show the performance of the proposed heuristic to plan hybrid driving-stepping locomotion. A video which shows additional footage of the experiments is available online\footnote{\url{https://www.ais.uni-bonn.de/videos/ICRA_2019_Klamt/}}.

\subsection{Abstraction Quality}
\label{sec:exp_abstr_quality}

\begin{figure}
	\centering
	\input{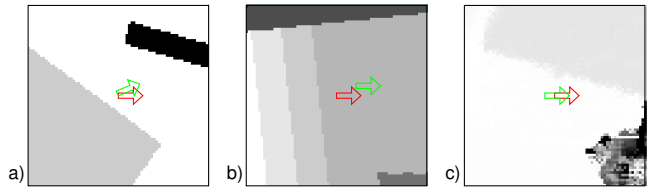} 
	\vspace{-0.7cm}
	\caption{Example tasks of the \emph{random} test set (a), \emph{simulated} test set (b), and \emph{real} test set (c). Red arrows show start states (not fed into the CNN), green arrows show goal states.}
	\label{fig:exp1_example_patches}
\end{figure}

The abstraction quality is evaluated on three data sets:
\begin{itemize}
	\item \emph{random}: We generate 200 random maps of each category resulting in a set of 1,124 maps with 24,728 tasks.
	\item \emph{simulated}: Height map patches of the desired size are cut out from height maps of simulated planning scenes. This set includes 77 maps with 1,694 tasks.
	\item \emph{real}:  Height map patches of the desired size are cut out from height maps that were generated from laser scanner measurements during real world experiments. This set includes 109 maps with 2,398 tasks.
\end{itemize}
We compared the performance to the manually tuned abstraction approach from our previous work.
Finally, costs for the tasks in the detailed representation $\mathcal{C}_\text{d}$ are stated as a base line.
We evaluate the \emph{feasibility} and \emph{costs} output.
A correct \emph{feasibility} assessment means that the abstract representation outputs the same feasibility value as the detailed representation.
Only if both representations assess a situation as feasible, \emph{costs} are considered and give an evaluation of the \emph{costs similarity} of the two representations.
\Cref{fig:exp1_example_patches} shows some example tasks.
The abstraction performance of the proposed CNN is shown in~\cref{tab:exp1_results}.

\begin{table}
	\centering
	\caption{Abstraction quality evaluation}
	\vspace{-0.2cm}
	\begin{tabular}[]{l|lll|}
		    & \textbf{\emph{random}} & \textbf{\emph{simulated}} & \textbf{\emph{real}} \\ \hline \hline 
		$\O$ $\mathcal{C}_\text{d}$ & 0.476 & 0.466 & 0.509 \\
		Std. dev.($\mathcal{C}_\text{d}$) & 0.222 & 0.202 & 0.236 \\ \hline
		\emph{feasibility} correct, CNN & 95.04\% & 96.69\% & 92.62\% \\
		$\O$ $\mathcal{C}_\text{a,CNN}$ & 0.453 & 0.469 & 0.446 \\
		$\O$ Error($\mathcal{C}_\text{a,CNN}$) & 0.027 & 0.013 & 0.081  \\ \hline
		\emph{feasibility} correct, man.tuned & 79.27\% & 65.35\% & 69.77\% \\
		$\O$ $\mathcal{C}_\text{a,man.tuned}$ & 0.435 & 0.402 & 0.429 \\
		$\O$ Error($\mathcal{C}_\text{a,man.tuned}$) & 0.057 & 0.021 & 0.103 \\ \hline
	\end{tabular}
	\label{tab:exp1_results}
	\vspace{-0.4cm}
\end{table}

The results indicate that the CNN \emph{feasibility} output is significantly better compared to the manually tuned abstraction.
While the latter has problems in simulated and real world robot environments, the CNN assesses a correct \emph{feasibility} for \textgreater\,92.62\% of the tasks throughout all test sets.

Regarding the \emph{costs} assessment, the average \emph{costs} error of the CNN is smaller compared to the manually tuned abstraction on all test sets.
The error is particular small when seen in relation to the large distribution of the base line costs.
While the error of the proposed CNN is \textless\,5.67\% of the absolute \emph{costs} on the \emph{random} and \emph{simulated} test sets, it is 15.9\% on the \emph{real} test set.
This might be explained by noisier sensor measurements which result in noisier height maps.

\subsection{Application to Planning}
\label{sec:exp_planning}

\begin{figure}
	\input{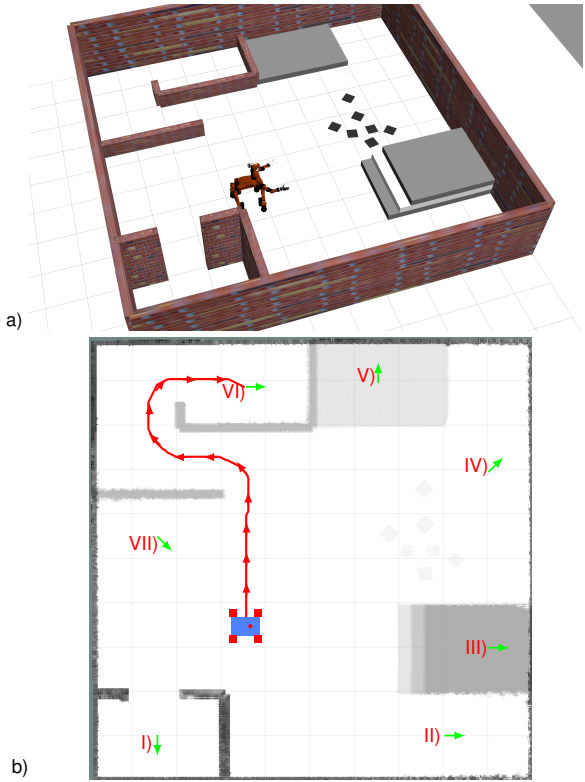}
	\vspace{-0.2cm}
	\caption{Locomotion planning experiment. a) Gazebo arena with Centauro. b) Height map with start state (blue/red) and goals (arrows): I) Behind a narrow door, II) next to stairs, III) on top of stairs, IV) behind some clutter, V) on a platform, VI) inside a labyrinth, and VII) behind the robot. The red path is the resulting path to VI with the proposed heuristic and $\mathcal{W}=1.25$.}
	\label{fig:robot_experiment_task}
	\vspace{-0.45cm}
\end{figure}

We designed a 10\,$\times$\,10\,m arena in Gazebo simulation which includes typical locomotion tasks for Centauro in search and rescue missions (\cref{fig:robot_experiment_task}).
Environment perception is realized through a continuously rotating Velodyne Puck 3D laser scanner with spherical field-of-view at the robot head.
Sensor measurements are processed to registered point clouds and used for localization using the method by Droeschel et al.~\cite{Droeschel2017104}.
Height maps are generated from these point clouds.
The used system is equipped with an Intel Core i7-8700K@3.70\,GHz, 64\,GB RAM and an NVidia GeForce GTX 1080Ti with 11\,GB memory.

For all abstract states, map patches are extracted and neighbors with costs are precomputed by the CNN.
This takes 239\,s, is only required once per map, and can be incrementally updated if parts of the map change.
The one-to-any Dijkstra search which starts from each goal state and generates the heuristic takes 0.049\,s in average.
We use the weighted-A* planner to plan paths to all goals while using the learned abstract representation as a heuristic.
We compare the planning performance to a geometric heuristic.
This combines Euclidean distances with rotational differences and is admissible.
Hence, when used with a weight $\mathcal{W}=1$, results are optimal.
Both heuristics are evaluated with multiple $\mathcal{W} \geq 1$ to also obtain fast, sub-optimal solutions.

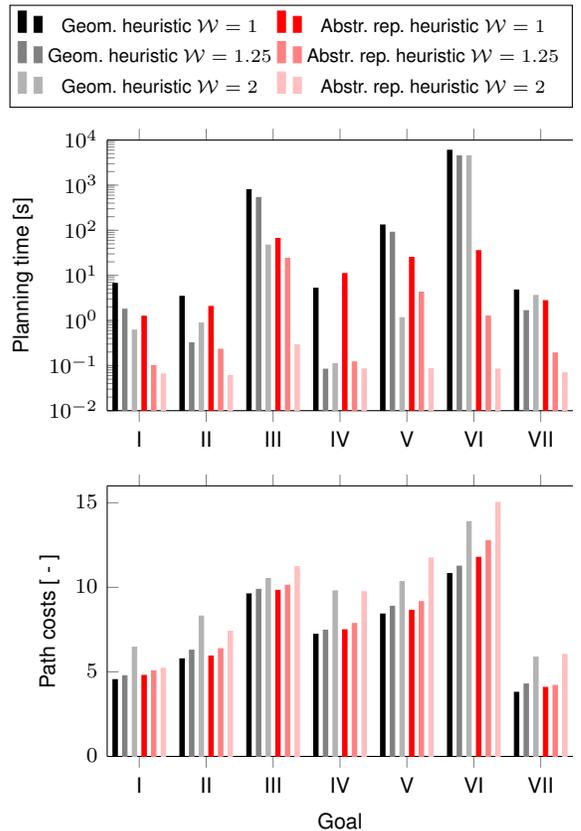
\begin{figure}
\centering
	\begin{tikzpicture}[
 	font=\sffamily\footnotesize,
    every node/.append style={text depth=.2ex},
	l/.style={font=\sffamily\scriptsize},
    b/.style={font=\bf},
]


\begin{groupplot}[group style = {group size = 1 by 2}]

\nextgroupplot[
			scaled ticks = false,
			ymin=0.01, ymax=10000,
			ymode=log,
			ytick={0.01, 0.1, 1, 10, 100, 1000, 10000},
			yticklabel style={/pgf/number format/fixed},
			log origin=infty,
			ylabel={Planning time [s]},
			width=0.9*\linewidth,
			height=0.6\linewidth,
			xlabel near ticks,
			ylabel near ticks,
			axis y line*=left,
			ybar,
			bar width=1.65pt,
			enlarge x limits=0.08,
			symbolic x coords={I, II, III, IV, V, VI, VII},
			legend style={at={(1,1.5)},font=\sffamily\scriptsize},
			legend columns=3,
			transpose legend
]
\addplot [color=black, fill] coordinates{(I,6.63) (II,3.42) (III,794)(IV, 5.18)(V, 130.9)(VI, 5932)(VII, 4.73)};
\addplot [color=black!50, fill] coordinates{(I,1.77) (II, 0.32) (III, 532)(IV, 0.083)(V, 90.47)(VI, 4470)(VII, 1.65)};
\addplot [color=black!30, fill] coordinates{(I,0.61) (II,0.88) (III, 46.89)(IV, 0.11)(V, 1.15)(VI, 4489)(VII, 3.59)};
\addplot [color=red, fill] coordinates{(I,1.24) (II,2.045) (III,65.94)(IV, 10.96)(V, 25.13)(VI, 35.27)(VII, 2.75)};
\addplot [color=red!50, fill] coordinates{(I,0.10) (II,0.23) (III,23.88)(IV, 0.12)(V, 4.25)(VI, 1.25)(VII, 0.19)};
\addplot [color=red!25, fill] coordinates{(I,0.065) (II,0.06) (III,0.29)(IV, 0.084)(V, 0.085)(VI, 0.083)(VII, 0.069)};
\legend{Geom. heuristic $\mathcal{W}=1$,Geom. heuristic $\mathcal{W}=1.25$,Geom. heuristic $\mathcal{W}=2$, Abstr. rep. heuristic $\mathcal{W}=1$,Abstr. rep. heuristic $\mathcal{W}=1.25$, Abstr. rep. heuristic $\mathcal{W}=2$}

\nextgroupplot[
			ymin=0, ymax=16,
			xlabel={Goal},
			ylabel={Path costs [ - ]},
			width=0.9*\linewidth,
			height=0.6\linewidth,
			xlabel near ticks,
			ylabel near ticks,
			axis y line*=left,
			ybar,
			bar width=1.65pt,
			enlarge x limits=0.08,
			symbolic x coords={I, II, III, IV, V, VI, VII},
			legend style={at={(1,1.3)},font=\sffamily\scriptsize},
]
\addplot [color=black, fill] coordinates{(I,4.54) (II,5.77) (III,9.61)(IV, 7.22)(V, 8.42)(VI, 10.81)(VII, 3.79)};
\addplot [color=black!50, fill] coordinates{(I,4.77) (II, 6.28) (III, 9.88)(IV, 7.46)(V, 8.88)(VI, 11.25)(VII, 4.28)};
\addplot [color=black!30, fill] coordinates{(I,6.46) (II,8.3) (III,10.52)(IV, 9.79)(V, 10.34)(VI, 13.87)(VII, 5.87)};
\addplot [color=red, fill] coordinates{(I,4.79) (II,5.93) (III, 9.82)(IV, 7.49)(V, 8.64)(VI, 11.77)(VII, 4.08)};
\addplot [color=red!50, fill] coordinates{(I,5.06) (II,6.37) (III, 10.12)(IV, 7.86)(V, 9.15)(VI, 12.75)(VII, 4.20)};
\addplot [color=red!25, fill] coordinates{(I,5.21) (II, 7.40) (III, 11.22)(IV,  9.74)(V, 11.73)(VI, 15.02)(VII, 6.03)};

\end{groupplot}

\end{tikzpicture}
	\vspace{-0.4cm}
	\caption{Planning times (including heuristic generation) and path costs for all goal states. The heuristic which is based on the learned abstract representation is compared to the geometric heuristic.}
	\label{fig:exp2_results}
\end{figure}

\Cref{fig:exp2_results} visualizes the planner performance for both heuristics and different $\mathcal{W}$.
Table~\ref{tab:speedup} summarizes the resulting speedup and cost increase compared to the optimal solution.
The results indicate that the proposed abstraction-based heuristic accelerates planning by multiple orders of magnitude while, in particular for $\mathcal{W}=1.25$, path costs stay comparable.
This significantly outperforms the geometric heuristic.
Especially for challenging tasks such as the stairs (III) and the labyrinth (VI), our heuristic was mandatory to obtain a solution in feasible time.
This can be explained by the fact that the geometric heuristic has no information about the environment and thus the planner may expand many states before considering expensive actions.
In contrast, the proposed abstraction-based heuristic uses its costs assessments to support the planner in its goal-directed behavior by including knowledge about the environment.

\begin{table}
	\centering
	\newcommand{\mc}[3]{\multicolumn{#1}{#2}{#3}}
	\caption{Heuristic performance}
	\vspace{-0.15cm}
	\begin{tabular}{l|lll|ll}
						& \mc{3}{c}{Abstract representation} & \mc{2}{c}{Geometric} \\
		$\mathcal{W}$		& 1.0 & 1.25 & 2.0 & 1.25 & 2.0 \\ \hline
		speedup factor  & 27.80 & 708.5 & 10,860 & 12.00 & 27.88 \\
		costs increase 	& +4.77\% & +10.5\% & +33.1\% & +6.07\% & +33,9\% 
	\end{tabular}
	\label{tab:speedup}
	\vspace{-0.4cm}
\end{table}


\section{Conclusion}

In this paper, we propose a 3-dimensional abstract representation for a high-dimensional locomotion planning problem which employs a CNN to learn the cost function. 
The CNN maps a local planning task, consisting of a map patch and goal state, to a costs assessment for this task.
We demonstrate how such an abstract representation can generate an informed heuristic for search-based high-dimensional planning.
Experiments show that such a heuristic accelerates planning by multiple orders of magnitude, especially for challenging tasks.
We further show that the learned representation outperforms a manually tuned abstract representation from previous work while eliminating tuning efforts.



\bibliographystyle{IEEEtranN}
\bibliography{references.bib} 

\addtolength{\textheight}{-12cm}   





\end{document}